\documentclass[letterpaper]{article} 
\usepackage{aaai2026}  
\usepackage{times}  
\usepackage{helvet}  
\usepackage{courier}  
\usepackage[hyphens]{url}  
\usepackage{graphicx} 
\usepackage{natbib}  
\usepackage{caption} 
\usepackage{algorithm}
\usepackage{algorithmic}
\setcitestyle{authoryear,open={(},close={)}}
\usepackage{subcaption}
\usepackage{booktabs}
\usepackage{multirow}
\usepackage{multicol}
\usepackage{amssymb}
\usepackage{placeins}
\usepackage{pifont}  
\newcommand{\xmark}{\ding{55}} 
\usepackage[table,xcdraw]{xcolor} 
\usepackage{enumitem}
\usepackage{wasysym}
\usepackage{amssymb}

\usepackage{amsmath}
\usepackage{adjustbox}
\usepackage{appendix}

\usepackage{newfloat}
\usepackage{listings}
\DeclareCaptionStyle{ruled}{labelfont=normalfont,labelsep=colon,strut=off} 
\floatstyle{ruled}
\newfloat{listing}{tb}{lst}{}
\floatname{listing}{Listing}
\title{\raisebox{-1mm}{\includegraphics[width=0.04\textwidth]{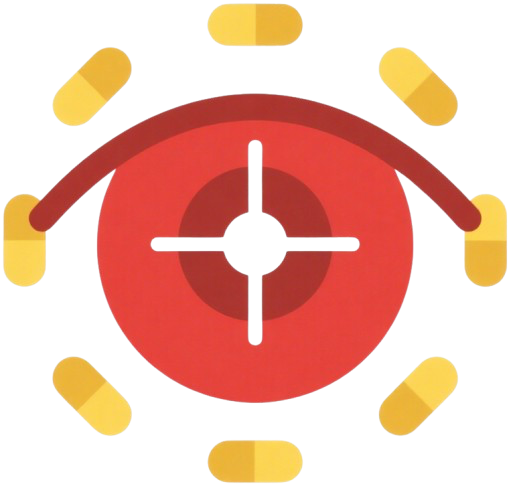}} MedEyes: Learning Dynamic Visual Focus for Medical Progressive Diagnosis}

\author{
    Chunzheng Zhu,
    Yangfang Lin,
    Shen Chen,
    Yijun Wang$^{*}$,
    Jianxin Lin\thanks{Corresponding authors.}
}
\affiliations{
    Hunan University\\
    Changsha, Hunan, China\\
    zhuchzh@hnu.edu.cn, lyfang123@hnu.edu.cn, cs05@hnu.edu.cn,\\
    wyjun@hnu.edu.cn, linjianxin@hnu.edu.cn
}

\usepackage{bibentry}

\begin{document}

\maketitle

\begin{abstract}

Accurate medical diagnosis often involves progressive visual focusing and iterative reasoning, characteristics commonly observed in clinical workflows. While recent vision-language models demonstrate promising chain-of-thought (CoT) reasoning capabilities via reinforcement learning with verifiable rewards (RLVR), their purely on-policy learning paradigm tends to reinforce superficially coherent but clinically inaccurate reasoning paths. We propose MedEyes, a novel reinforcement learning framework that dynamically models clinician-style diagnostic reasoning by progressively attending to and interpreting relevant medical image regions. By incorporating off-policy expert guidance, MedEyes converts expert visual search trajectories into structured external behavioral signals, guiding the model toward clinically aligned visual reasoning. We design the Gaze-guided Reasoning Navigator (GRN) to emulate the diagnostic process through a dual-mode exploration strategy, scanning for systematic abnormality localization and drilling for detailed regional analysis. To balance expert imitation and autonomous discovery, we introduce the Confidence Value Sampler (CVS), which employs nucleus sampling and adaptive termination to create diverse yet credible exploration paths. Finally, the dual-stream GRPO optimization framework decouples on-policy and off-policy learning signals, mitigating reward assimilation and entropy collapse. Experiments demonstrate that MedEyes achieves an average performance improvement of +8.5pp across multiple medical VQA benchmarks, validating MedEyes's potential in building trustworthy medical AI systems. Code is available at \url{https://github.com/zhcz328/MedEyes}.

\end{abstract}

%

\begin{figure}[t]

  \includegraphics[width=0.497\textwidth]{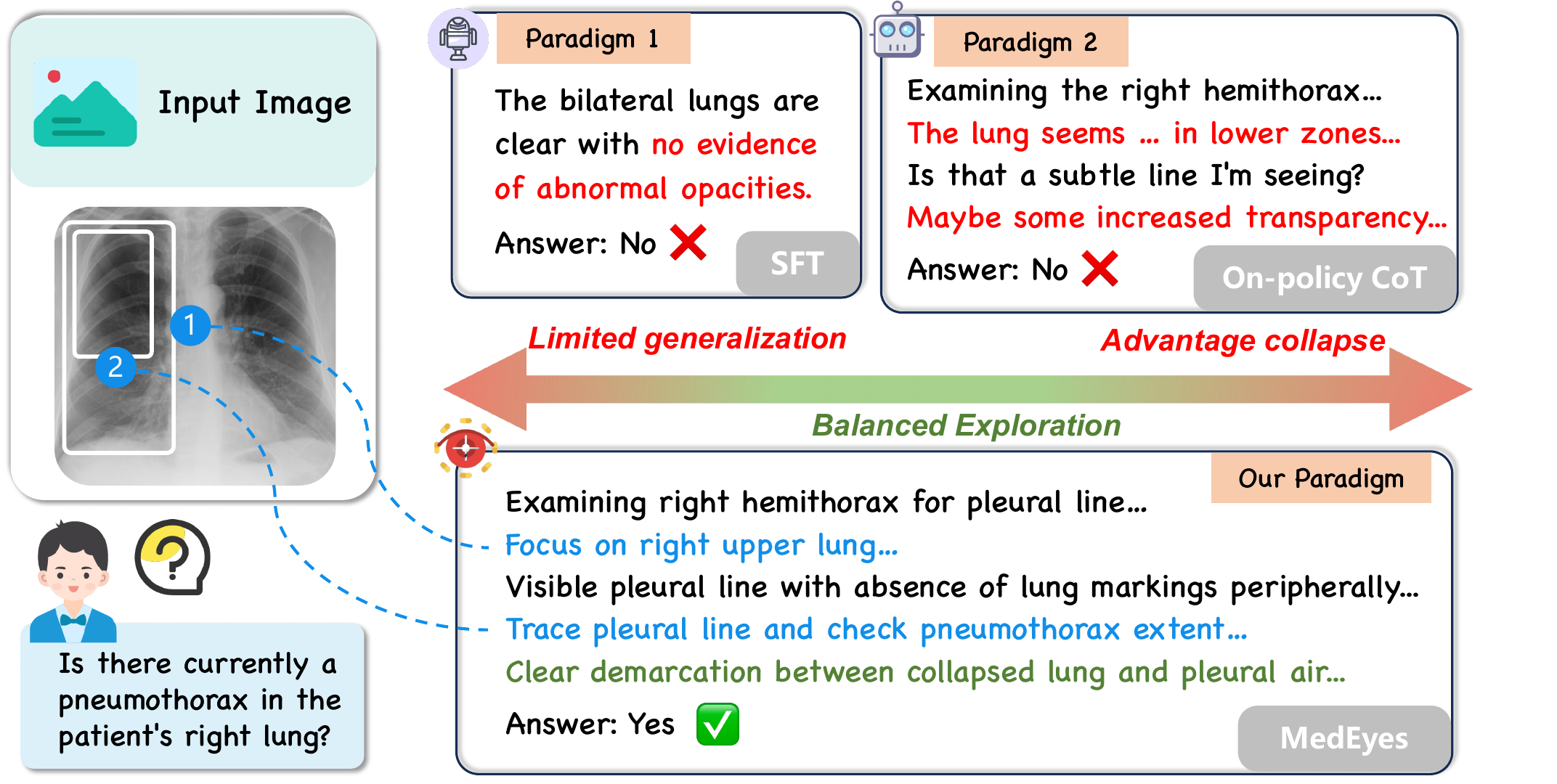}
     \caption{Comparison of medical CoT training paradigms: SFT produces overly generic responses that miss critical findings; on-policy CoT allows exploration but suffers from advantage collapse leading to incorrect reasoning; MedEyes achieves accurate pneumothorax identification through systematic visual grounding and targeted regional analysis.}
     \label{fig:motivation}

\end{figure}

\section{Introduction}

Recent breakthroughs in medical vision-language models (VLMs), such as RadFM~\cite{wu2023towards}, PathChat~\cite{lu2024multimodal}, and MMed-RAG~\cite{xia2025mmed}, have demonstrated exceptional capabilities in medical question answering and report generation tasks. These models leverage large-scale multimodal pretraining to support complex diagnostic reasoning and clinical decision-making. Beyond static diagnosis, recent large multi-step reasoning models (LRMs) have demonstrated promising capabilities in Chain-of-Thought (CoT) reasoning and self-reflective inference, particularly in text-based medical image interpretation tasks~\cite{lai2025med, pan2025medvlm}. More recently, the emergence of visual CoT methods~\cite{openai2025thinkingwithimages, zheng2025deepeyes, fan2025grit} has opened new directions for progressive, vision-grounded reasoning, bridging the gap between textual logic and spatial evidence.

Supervised fine-tuning (SFT) captures task-specific knowledge through large-scale CoT data but often overfits to memorized trajectories, compromising generalization and faithfulness in unseen clinical scenarios~\cite{yu2025finemedlm, zhuang2025wingpt}. While text-based CoT and reinforcement learning with verifiable rewards (RLVR) improve interpretability and accuracy~\cite{lai2025med, pan2025medvlm}, they primarily operate in the textual domain. This lack of explicit grounding between reasoning steps and visual evidence triggers information loss and visual hallucinations in complex imaging tasks. As shown in the pneumothorax example (Fig.~\ref{fig:motivation}), SFT models yield vague responses, whereas on-policy CoT reasoning frequently suffers from "advantage collapse," generating plausible but substantively incorrect paths that lead to erroneous conclusions.

These limitations raise a critical question in medical visual reasoning: \textit{How can we enable models to acquire the progressive visual focus and iterative diagnostic refinement that characterize expert clinical workflows?}

Accurate medical diagnosis relies on expert-level reasoning supported by high-quality exploration trajectories. However, models initialized with limited capabilities often fall into local optima, resulting in repetitive, low-quality reasoning cycles, which we refer to as cognitive traps. Off-policy expert trajectories can act as cognitive anchors, guiding effective diagnostic exploration. Yet, naive behavior cloning tends to mimic action sequences without capturing their underlying reasoning logic, while overly exploring may deviate from clinically valid cognitive structures. Bridging this gap requires a hybrid learning paradigm that combines expert-guided supervision with self-driven exploration. Such a framework should support not only behavioral imitation but also knowledge-level generalization, enabling models to internalize external clinical expertise as intrinsic reasoning skills. Moreover, when reasoning steps are explicitly grounded to visual regions, the model can achieve a natural alignment between ``precise observation'' and ``structured reasoning'', thereby establishing a consistent mapping between image evidence and diagnostic descriptions.


This work introduces MedEyes, a hybrid reinforcement learning framework that captures dynamic clinician-style attention patterns during diagnostic processes through medical CoT reasoning. Our approach leverages structured off-policy expert trajectories as cognitive anchors, enabling models to internalize expert diagnostic behaviors while preventing policy collapse. Specifically, we design the Gaze-guided Reasoning Navigator (GRN), implementing dual-mode exploration: a scanning mode for identifying candidate abnormal regions, and a drilling mode for conducting focused pathological analysis. This mirrors human diagnostic workflows as captured in eye-tracking studies~\cite{brunye2019review,castner2020deep,sultana2024seeing}. To balance expert imitation with autonomous exploration, the Confidence Value Sampler (CVS) generates high-quality reasoning trajectories via nucleus sampling and dynamically adjusts exploration depth based on confidence feedback. Finally, we develop a dual-stream variant of Group Relative Policy Optimization (GRPO)~\cite{shao2024deepseekmath} that decouples training signals from on-policy and off-policy sources. This design mitigates reward assimilation by preventing expert trajectories from dominating autonomous learning signals and avoids entropy collapse that constrains exploration and impairs generalization to novel diagnostic scenarios.

We evaluate MedEyes on five established medical visual question answering benchmarks covering diverse imaging modalities, ranging from radiological analysis to histopathological examination. Our experiments reveal that the proposed framework successfully enables training of initially weak models, which prove intractable under pure on policy reinforcement learning, while exhibiting clinically aligned progressive visual attention and iterative diagnostic reasoning. These findings establish MedEyes as a transformative paradigm for developing interpretable medical AI systems with generalizable visual reasoning capabilities. 
The main contributions of this paper can be summarized as follows:

\begin{itemize}
\item We propose MedEyes, a dynamic focusing multi-round reasoning RL framework, which breaks through the limitations of traditional medical post-training by introducing structured off-policy expert trajectories.

\item We design a collaborative mechanism between the Gaze-guided Reasoning Navigator (GRN) and Confidence Value Sampler (CVS), where the former reproduces diagnostic workflows through scanning-drilling dual-mode strategies, and the latter constructs a diverse and high-quality off-policy trajectory library.

\item We use dual-stream GRPO optimization architecture to address reward assimilation and entropy collapse issues by isolating on-policy and off-policy learning components, achieving a balance between expert-level diagnostic pattern learning and task adaptability.

\item Comprehensive validation on five medical visual question answering benchmarks shows that this method not only significantly outperforms existing methods, but achieves breakthroughs in clinical interpretability and visual localization accuracy, providing a new technical pathway for building trustworthy medical AI systems.
\end{itemize}

\begin{figure*}[t]
  \includegraphics[width=0.995\textwidth]{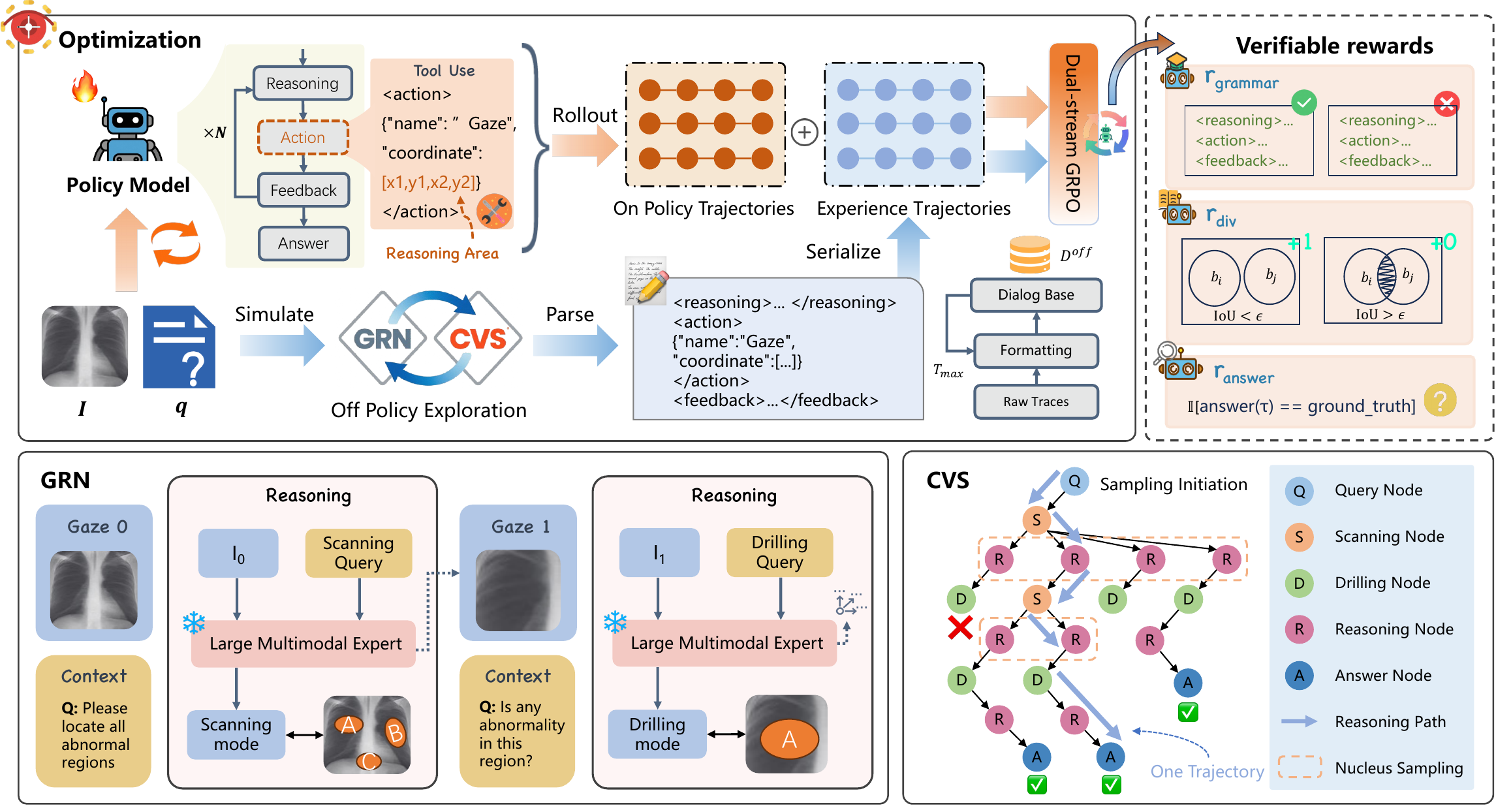}
  \caption{Overview of MedEyes. We first generate structured off-policy expert trajectories through the Gaze-guided Reasoning Navigator (GRN) and Confidence Value Sampler (CVS) to explore expert reasoning patterns, then combine these with on-policy rollouts from the policy model. The unified trajectory is subsequently  for optimization via dual-stream GRPO with multi-component verifiable rewards to enhance the model's intrinsic grounding reasoning capability for medical visual understanding.
}
\end{figure*}

\section{Methodology}
\subsection{Framework Overview}
MedEyes presents a mixed-policy RL framework leveraging dual-stream GRPO optimization to overcome the cognitive traps and policy collapse endemic to SFT and pure on-policy paradigms through structured off-policy expert trajectory integration. The architecture contains two synergistic components: (1) an on-policy exploration stream, where the policy $\pi_\theta$ autonomously samples diagnostic trajectories; and (2) an off-policy guidance stream, which constructs expert trajectories via the Gaze-guided Reasoning Navigator (GRN) and Confidence Value Sampler (CVS), establishing cognitive anchors that help the agent generalize beyond initialization and support progressive refinement of diagnostic reasoning.

\subsubsection{Gaze-guided Multi-round Reasoning}

Medical visual reasoning is formalized as a Markov decision process enabling progressive diagnostic attention refinement through reinforcement learning mechanisms. Given medical image $I$ and clinical query $q$, policy $\pi_\theta$ generates diagnostic trajectory $\tau = [n_1, n_2, ..., n_T, a]$ where each reasoning step $n_t = \langle s_t, \mathcal{G}_t \rangle$ encapsulates textual cognition $s_t$ and visual grounding $\mathcal{G}_t = \{(x_{i,1}, y_{i,1}, x_{i,2}, y_{i,2})\}$, culminating in diagnostic answer $a$. The reasoning process manifests as a factorized probability distribution over trajectory space:
\begin{equation}
\normalsize
\scalebox{0.97}{$
\begin{aligned}
\pi_\theta(\tau \mid I, q) = \prod_{t=1}^T \pi_\theta(n_t \mid I, q, n_{<t}) \cdot \pi_\theta(a \mid I, q, n_{\leq T}),
\end{aligned}$}
\end{equation}
where policy $\pi_\theta$ parameterized by $\theta$ maximizes expected rewards across hybrid trajectory distributions, acquiring clinically aligned progressive attention mechanisms that circumvent initialization bottlenecks.
Progressive attention materializes through structured trajectory sequences embodying clinical diagnostic workflows:
$\tau = \left[\prod_{i=1}^{T} \left(\theta_i \circ \alpha_i \circ o_i\right)\right] \circ \phi$,
where concatenation operator $\circ$ links reasoning step $\theta_i$, tool invocation $\alpha_i$, and observational feedback $o_i$, terminating with final answer $\phi$. Each reasoning-action-perception triplet $(\theta_i, \alpha_i, o_i)$ instantiates expert diagnostic protocols established in radiological interpretation studies. This architecture facilitates continuous visual-cognitive interaction, achieving diagnostic convergence through iterative grounding and reasoning refinement that mirrors expert clinician behavioral patterns documented in clinical practice.
\subsection{Off-Policy Expert Trajectory Generation}
\label{sec:offpolicy}

To enhance policy learning from external behaviors, expert search protocols are systematically transformed into structured learning signals via the GRN for dual-mode exploration and the CVS for trajectory sampling diversification.

\subsubsection{Gaze-guided Reasoning Navigator (GRN)}
\label{sec:grn}

Aligned with clinical eye-tracking protocols~\cite{brunye2019review}, GRN maintains ternary attention state $\psi_t = (\mathcal{R}_t, \mathcal{C}_t, \mathcal{F}_t)$ orchestrating visual exploration dynamics, where $\mathcal{R}_t$ constitutes candidate region sets generated via large-scale multimodal expert consultation through region-level VQA queries and $\langle \text{SEG}\rangle$ segmentation prompts, $\mathcal{C}_t$ contains corresponding confidence distributions, and $\mathcal{F}_t \in \{\text{global}, \text{local}\}$ specifies exploration modes. The state evolution follows:
\begin{equation}
	\psi_{t+1} = \mathcal{T}(\psi_t, a_t, o_t),
\end{equation}
where $a_t$ is the action taken and $o_t$ is the resulting observation. This transition controls how the model updates its attention state based on new visual diagnostic evidence.

\textbf{Dual-mode Exploration Strategy} The state transition $\mathcal{T}$ operates through two complementary exploration modes that reflect expert physicians' visual search patterns:

\textit{Scanning Mode}: When $\mathcal{F}_t = \text{global}$, GRN submits the prompt ``Please locate all abnormal regions in the image $\langle \text{SEG}\rangle$" to the expert model, generating comprehensive candidate regions $\mathcal{R}_{t+1} = \{r_1, r_2, ..., r_n\}$ and populating the corresponding confidence scores $\mathcal{C}_{t+1}$ for the next state.

\textit{Drilling Mode}: When $\mathcal{F}_t = \text{local}$, GRN performs targeted analysis of the candidate regions from the current state $\mathcal{C}_t$ using the targeted diagnostic query, such as: ``Please analyze the abnormality in region $\langle\text{region}\rangle r_i \langle/\text{region}\rangle$ $\langle\text{SEG}\rangle$". This detailed pathological analysis produces refined confidence scores that update $\mathcal{C}_{t+1}$ for the specific analyzed region $r_i$. The confidence evolution between consecutive states determines the subsequent exploration mode $\mathcal{F}_{t+1}$:
\begin{equation}
	\Delta c = \frac{c_{t+1}(r_i) - c_t(r_i)}{c_t(r_i) + \epsilon},
\end{equation}
where $c_t(r_i) \in \mathcal{C}_t$ and $c_{t+1}(r_i) \in \mathcal{C}_{t+1}$ denote prior and refined confidence scores for region $r_i$, respectively, and $\epsilon$ is a stability constant. If $\Delta c \geq \delta$, drilling mode persists. Otherwise, scanning mode resumes for broader exploration.

\subsubsection{Confidence Value Sampler (CVS)}
\label{sec:cvs}
To obtain diverse yet credible expert behaviors and capture diagnostic reasoning patterns, CVS applies nucleus sampling~\cite{holtzman2019curious} to GRN's multi-round trajectories, generating multiple variable-length exploration paths. At each decision step $t$, the sampler selects from the top-$p_{0}$ confidence regions:
\begin{equation}
\mathcal{P}_{\text{nucleus}} = \left\{ a_i : \sum_{j=1}^{i} P(a_j \mid \psi_t) \leq p_0 \right\},
\end{equation}
where actions are ranked in descending probability order by $P(a_j \mid \psi_t)$ based on state $\psi_t$, with candidate region selection for $\mathcal{C}_{t+1}$ within the scanning-drilling exploration. Conditioned on GRN's region proposals, the CVS generates distinct trajectories $N_{\text{expert}}=\{\tau_1^{\text{expert}}, \tau_2^{\text{expert}}, ..., \tau_{N_{\text{expert}}}^{\text{expert}}\}$. Each trajectory sampling terminates upon satisfying convergence criteria: local confidence exceeds threshold $\xi$ indicating diagnostic certainty, or maximum length $T_{\max}$ is reached.


\textbf{Trajectory Parsing and Serialization.} The raw visual exploration traces are then systematically parsed into structured dialog sequences through sequential decomposition. Each reasoning cycle is formatted using standardized tags: \texttt{<reasoning>...</reasoning>} for cognitive steps, followed by either \texttt{<action>{"name": "Gaze", "coordinate": [x1,y1,x2,y2]}</action>} for continued grounding operations with subsequent \texttt{<feedback>...</feedback>} containing coordinate-centered visual crops resized to target resolution, or direct termination with \texttt{<answer>...</answer>}. The serialization process converts these parsed trajectories into the multi-round format $(\theta_i, \alpha_i, o_i)$ while preserving temporal dependencies and spatial grounding relationships.

This mechanism captures diagnostic complexity variation across cases, where simple abnormalities enable immediate detection while complex scenarios demand extensive multi-region exploration. Each trajectory $\tau^{\text{expert}}_k = \{(s_1, \mathcal{G}_1), ..., (s_{T^*_k}, \mathcal{G}_{T^*_k}), a_k\}$ represents a complete reasoning path with adaptive length $T^*_k \leq T_{\max}$ determined by visual complexity rather than fixed templates. These structurally consistent yet content-diverse trajectories make up the off-policy replay buffer $\mathcal{D}^{\text{off}}$ for subsequent training.

\begin{table*}[t]
\centering
\renewcommand{\arraystretch}{1}
\small
\adjustbox{width=0.94\textwidth}{
\begin{tabular}{l|c>{\centering\arraybackslash}p{1.25cm}>{\centering\arraybackslash}p{1.25cm}c>{\centering\arraybackslash}p{1.25cm}|>{\centering\arraybackslash}p{1.3cm}|>{\centering\arraybackslash}p{1.3cm}}
\toprule[1pt]
\textbf{Method} & \textbf{VQA-RAD} & \textbf{SLAKE} & \textbf{PathVQA} & \textbf{PMC-VQA} & \textbf{MMMU*} & \textbf{Average} & $\mathbf{\Delta}$ \textbf{(\%)} \\
\midrule[0.85pt]
\multicolumn{8}{l}{\textit{General Vision-Language Models}} \\
Qwen2.5-VL-3B~\cite{Bai2025Qwen2.5-VL} & 47.3 & 56.2 & 55.2 & 37.5 & 34.3 & 46.1 & -19.8 \\
GPT-4o~\cite{hurst2024gpt} & 54.2 & 50.1 & \text{59.2} & 40.8 & -- & 51.1 & -14.8 \\
InternVL-2~\cite{chen2024internvl} & 58.8 & 62.7 & 45.8 & 38.4 & \text{52.7} & 51.7 & -14.2 \\
\midrule[0.85pt]
\multicolumn{8}{l}{\textit{Medical-Specific Models}} \\
RadFM~\cite{wu2023towards} & 50.6 & 34.6 & 38.7 & 25.9 & 27.0 & 35.4 & -30.5 \\
MedVInT~\cite{zhang2023pmc} & 45.4 & 43.5 & 54.7 & 23.3 & 28.3 & 39.0 & -26.9 \\
LLaVA-Med~\cite{li2023llava} & 51.4 & 48.6 & 56.8 & 24.7 & 36.9 & 43.7 & -22.2 \\
Med-Flamingo~\cite{moor2023med} & 55.8 & 59.6 & 40.7 & 34.7 & 47.5 & 47.7 & -18.2 \\
GMAI-VL~\cite{li2024gmai} & \text{64.6} & \text{71.9} & 47.2 & \text{52.3} & 51.2 & \text{57.4} & -8.5 \\
\midrule[0.85pt]
\multicolumn{8}{l}{\textit{Reinforcement Learning Methods}} \\
GRIT$\dagger$~\cite{fan2025grit} & 54.3 & 57.1 & 43.5 & 42.3 & 49.5 & 49.3 & -16.6 \\
DeepEyes$\dagger$~\cite{zheng2025deepeyes} & 56.4 & 59.7 & 42.3 & 45.2 & 49.1 & 50.5 & -15.4 \\
Med-R1~\cite{lai2025med} & 55.9 & 55.1 & 53.3 & 45.8 & 32.7 & 48.5 & -17.4 \\
MedVLM-R1~\cite{pan2025medvlm} & 61.4 & 65.9 & 55.2 & 44.8 & 35.5 & 52.5 & -13.4 \\
\midrule[0.85pt]
\textbf{MedEyes (Ours)} & \textbf{70.7} & \textbf{79.1} & \textbf{64.8} & \textbf{55.3} & \textbf{59.7} & \textbf{65.9} & -- \\
\bottomrule[1pt]
\end{tabular}
}
\caption{Comprehensive medical VQA benchmark performance comparison across five widely-used common open medical datasets. Best results are marked in \text{bold}. $\Delta$ indicates the performance gap (\%) compared to our method.}
\label{tab:main_results}
\end{table*}

\subsection{Dual-stream GRPO Reinforcement Learning}

Standard GRPO~\cite{shao2024deepseekmath} employs a single advantage normalization across all trajectories, while our approach decouples the advantage computation for on-policy and off-policy data, avoiding gradient dominance that undermines mixed-policy training effectiveness.

\subsubsection{Verifiable Reward Function}
\label{sec:reward}

The composite reward function $R(\tau) = \lambda_{\text{acc}} \cdot r_{\text{acc}}(\tau) + \lambda_{\text{grammar}} \cdot r_{\text{grammar}}(\tau) + \lambda_{\text{div}} \cdot r_{\text{div}}(\tau)$ comprises three components that address different aspects of medical visual reasoning: diagnostic accuracy, structural correctness, and exploration diversity.

\textbf{Accuracy Reward} Evaluates diagnostic correctness by comparing reasoning-generated answers with ground truth: 
\begin{equation}
r_{\text{acc}}(\tau) = \mathbb{I}[\text{answer}(\tau) = \text{ground\_truth}],
\end{equation}
where $\text{answer}(\tau)$ denotes the final answer extracted from the \texttt{<answer>...</answer>} tags from trajectory $\tau$, and $\mathbb{I}(.)$ is an indicator function.

\textbf{Grammar Reward} Ensures multi-round reasoning structure correctness through format validation:
\begin{equation}
\small
r_{\text{grammar}}(\tau) = \prod_{i=1}^{T} \mathbb{I}[W(\theta_i, \alpha_i, o_i)] \cdot \mathbb{I}[E(\tau)],
\end{equation}
where $W(\theta_i, \alpha_i, o_i)$ validates each reasoning cycle format following the standardized tag structure defined in Section \ref{sec:cvs}. The grammar reward is binary: $r_{\text{grammar}}(\tau) = 1$ if and only if all format validations succeed, and 0 otherwise, ensuring strict adherence to grammatical correctness.

\textbf{Diversity Reward}  Incentivizes comprehensive multi-region visual exploration:
\begin{equation}
\small
r_{\text{div}}(\tau) = \min\left(1, \frac{|\mathcal{U}(\tau)|}{n}\right) + \frac{1}{\binom{|\mathcal{U}(\tau)|}{2}} \sum_{i < j} \mathbb{I}[\text{IoU}(b_i, b_j) < \epsilon],
\end{equation}
where $\mathcal{U}(\tau)$ denotes unique regions explored in trajectory $\tau$, $n$ matches GRN's region count limit, and $b_i$, $b_j$ are bounding boxes from different tool calls. The first term rewards multiple distinct regions, while the second term encourages spatial diversity through sufficiently separated regions.

\subsubsection{Optimization Objective}
\label{sec:hybrid_grpo}

We propose mixed-policy optimization that achieves progressive evolution from expert imitation to autonomous discovery by decoupling advantage normalization for on-policy exploration and off-policy guidance. Policy parameters $\theta$ are updated by maximizing:
\begin{equation}
\small
\scalebox{0.94}{$
\begin{aligned}
\mathcal{J}(\theta) = \frac{1}{N}\sum_{i=1}^{N} \frac{1}{|\tau_i|}  \sum_{t=1}^{|\tau_i|}
    \min\!\bigl(\rho_{i,t}^{\theta} A_i,\; \mathrm{clip}(\rho_{i,t}^{\theta}, 1-\epsilon, 1+\epsilon)\,A_i \bigr),
\end{aligned}$}
\end{equation}
where $N = N_{\text{on}} + N_{\text{off}}$ denotes total trajectory count. In our hybrid learning paradigm, we employ source-adaptive importance ratio computation: for on-policy trajectories, $\rho_i^{\theta} = \pi_\theta(\tau_i \mid I,q) / \pi_{\theta_{\text{old}}}(\tau_i \mid I,q)$ measures policy evolution from the previous iteration, while for off-policy expert trajectories, $\rho_i^{\theta} = \pi_\theta(\tau_i \mid I,q) / \pi_{\text{expert}}(\tau_i \mid I,q)$ where $\pi_{\text{expert}}(\tau_i \mid I,q)=1$ represents the expert trajectory generation policy from CVS with nucleus sampling. This dual-stream design maintains gradient stability across heterogeneous trajectory sources while effectively mitigating potential distribution shifts between the respective signals.

\textbf{Advantage Decoupling Mechanism} We employ source-specific advantage normalization rather than unified normalization across all trajectories:
\begin{equation}
\small
\scalebox{0.99}{$
\begin{aligned}
A_i = \frac{R(\tau_i) - \mu^{s(i)}}{\sigma^{s(i)} + \varepsilon}, \quad \text{where } s(i) = \begin{cases}
\text{on}, & \text{if } \tau_i \in \mathcal{D}^{\text{on}} \\
\text{off}, & \text{if } \tau_i \in \mathcal{D}^{\text{off}}
\end{cases},
\end{aligned}$}
\end{equation}
where $\mathcal{D}^{\text{on}}$ represents on-policy trajectories and $\mathcal{D}^{\text{off}}$ represents off-policy expert trajectories. The statistics $\mu^{\text{on/off}} = \mathbb{E}_{\tau \sim \mathcal{D}^{\text{on/off}}}[R(\tau)]$ and $\sigma^{\text{on/off}} = \sqrt{\text{Var}_{\tau \sim \mathcal{D}^{\text{on/off}}}[R(\tau)]}$ are computed independently on their respective data distributions. This decoupling strategy prevents expert trajectories from overshadowing on-policy advantages, avoiding gradient dominance that undermines autonomous learning. Separate normalization maintains distinct learning signals, enabling effective knowledge transfer while preserving model's reasoning adaptability to novel medical cases.

\section{Experiments}

\subsection{Experimental Setup}

\paragraph{Datasets} We conduct comprehensive experiments on five widely adopted medical visual question answering benchmarks to evaluate the effectiveness of our method, including VQA-RAD~\cite{lau2018dataset}, SLAKE~\cite{liu2021slake}, PathVQA~\cite{he2020pathvqa}, PMC-VQA~\cite{zhang2023pmc}, and MMMU*~\cite{yue2024mmmu} (* indicates the Health\&Medicine subset). These datasets cover various medical imaging modalities, such as radiology (CT, MRI, X-ray), pathological slides, and multimodal medical scenarios. More details about the training datasets and statistics are provided in the supplementary materials.

\paragraph{Baselines }We compare MedEyes against three categories of comprehensive baselines: (1) \textit{General vision-language models}, including GPT-4o~\cite{hurst2024gpt}, Gemini-Pro~\cite{team2023gemini}, Qwen2.5-VL-3B~\cite{Bai2025Qwen2.5-VL}, and InternVL-2~\cite{chen2024internvl}, which we use for their thinking ability to perform reasoning; (2) \textit{Medical-specific VLMs}, including LLaVA-Med~\cite{li2023llava}, MedVInT~\cite{zhang2023pmc}, Med-Flamingo~\cite{moor2023med}, RadFM~\cite{wu2023towards}, GMAI-VL~\cite{li2024gmai}, which leverage multimodal understanding for visual question answering; (3) \textit{Reinforcement learning enhanced methods}, including visual CoT methods GRIT$\dagger$~\cite{fan2025grit} and DeepEyes$\dagger$~\cite{zheng2025deepeyes} ($\dagger$ denotes using the same dataset adapted to medical domains), and Med-R1~\cite{lai2025med} and MedVLM-R1~\cite{pan2025medvlm}, which leverage GRPO to generate explicit textual reasoning processes for enhanced diagnostic interpretability and reliability.

\paragraph{Implementation Details} Our framework builds upon  Qwen2.5‑VL‑3B~\cite{Bai2025Qwen2.5-VL}, leveraging its strong multimodal capabilities. We adopt MedPLIB~\cite{huang2025towards} as the sole visual expert to maximize task alignment. Within each reasoning step, the GRN processes $n{=}5$ candidate regions while maintaining established visual grounding formats. Mode transitions employ adaptive thresholds $\delta{=}0.15$ with stability constant $\epsilon{=}1{\times}10^{-6}$. CVS implements nucleus sampling ($p_0{=}0.9$) with termination when the confidence exceeds $\xi{=}0.85$ or reaches the maximum length $T_{\max}{=}4$, with max generation per turn set to 1024 tokens. This process generates $N_{\text{expert}}{=}6$ expert trajectories to form our off-policy replay buffer. 
The visual encoder processes images at a resolution of $336\times336$ with a patch size of 14. The dual-stream GRPO training process uses carefully balanced reward weights: $\lambda_{\text{acc}}{=}0.7$, $\lambda_{\text{grammar}}{=}0.2$, and $\lambda_{\text{div}}{=}0.1$. We optimized the model using AdamW with a learning rate of $1\times10^{-6}$, and conducted training over three epochs on six NVIDIA GeForce RTX 3090 GPUs. The training runs for 80 iterations with a rollout batch size of 98, generating 8 rollouts per prompt to ensure adequate exploration.

\begin{figure}[t]
  \includegraphics[width=0.49\textwidth]{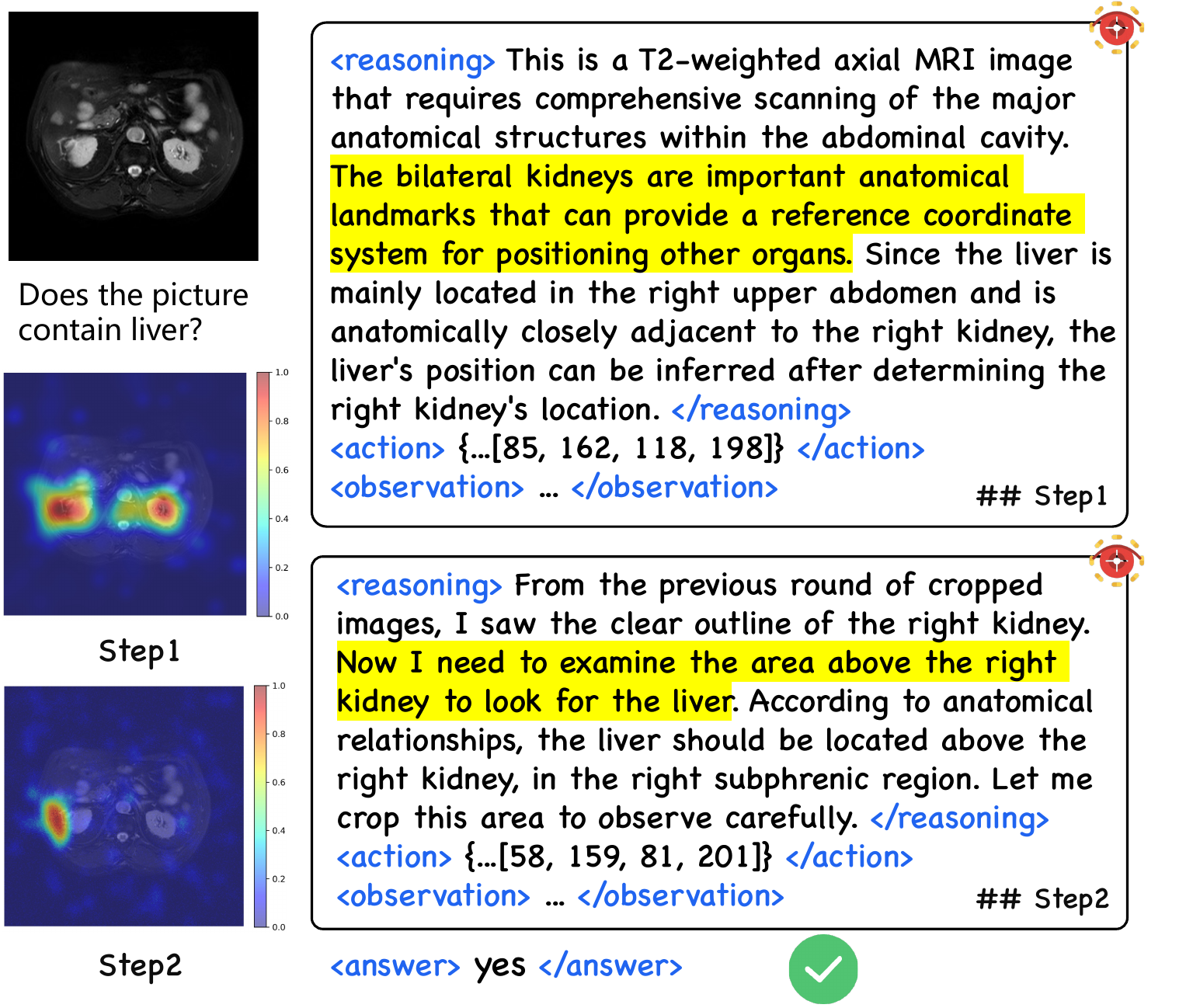}
	\caption{Diagnostic chain-of-thought example of MedEyes. Step 1 identifies bilateral kidneys as anatomical landmarks, followed by targeted liver in step 2. Heatmaps illustrate the progressive refinement process of visual attention.}
 \label{fig:case_analysis}
\end{figure}

\subsection{Main Results}

Table~\ref{tab:main_results} shows the comprehensive evaluation results of MedEyes across five medical VQA benchmarks. MedEyes achieves state-of-the-art performance with an average accuracy of 65.9, substantially outperforming all baseline categories. Specifically, it surpasses the best medical-specific model GMAI-VL by 8.5\% and the strongest reinforcement learning method MedVLM-R1 by 13.4\%. The superior performance demonstrates that vision-grounded chain-of-thought reasoning significantly outperforms purely textual approaches by establishing explicit visual-textual connections. On PathVQA, MedEyes achieves 64.8 accuracy, validating our dual-mode exploration strategy. On MMMU (Health \& Medicine), MedEyes consistently exceeds all baselines, confirming the efficacy of dual-stream GRPO optimization. These results validate that off-policy expert trajectory guidance with interleaved visual-textual reasoning successfully overcomes pure on-policy learning limitations.

\subsection{In-Depth Case Analysis}
As illustrated in Fig.~\ref{fig:case_analysis}, the T2-weighted MRI liver detection case demonstrates MedEyes's progressive diagnostic reasoning. The framework initiates systematic anatomical localization through bilateral kidney identification as spatial reference coordinates, subsequently executing targeted exploration of the region adjacent to the right kidney. Cross-attention weight visualizations from the vision-language transformer adapter directly map reasoning tokens to corresponding image regions, revealing stepwise attention refinement from diffuse activation patterns to diagnostically salient areas. This progression validates the framework's capacity to internalize expert-level visual search hierarchies, wherein diagnostic confidence emerges through systematic evidence synthesis rather than pattern recognition heuristics.

\subsection{Training Dynamics Analysis}

\begin{figure}[t]
  \includegraphics[width=0.47\textwidth]{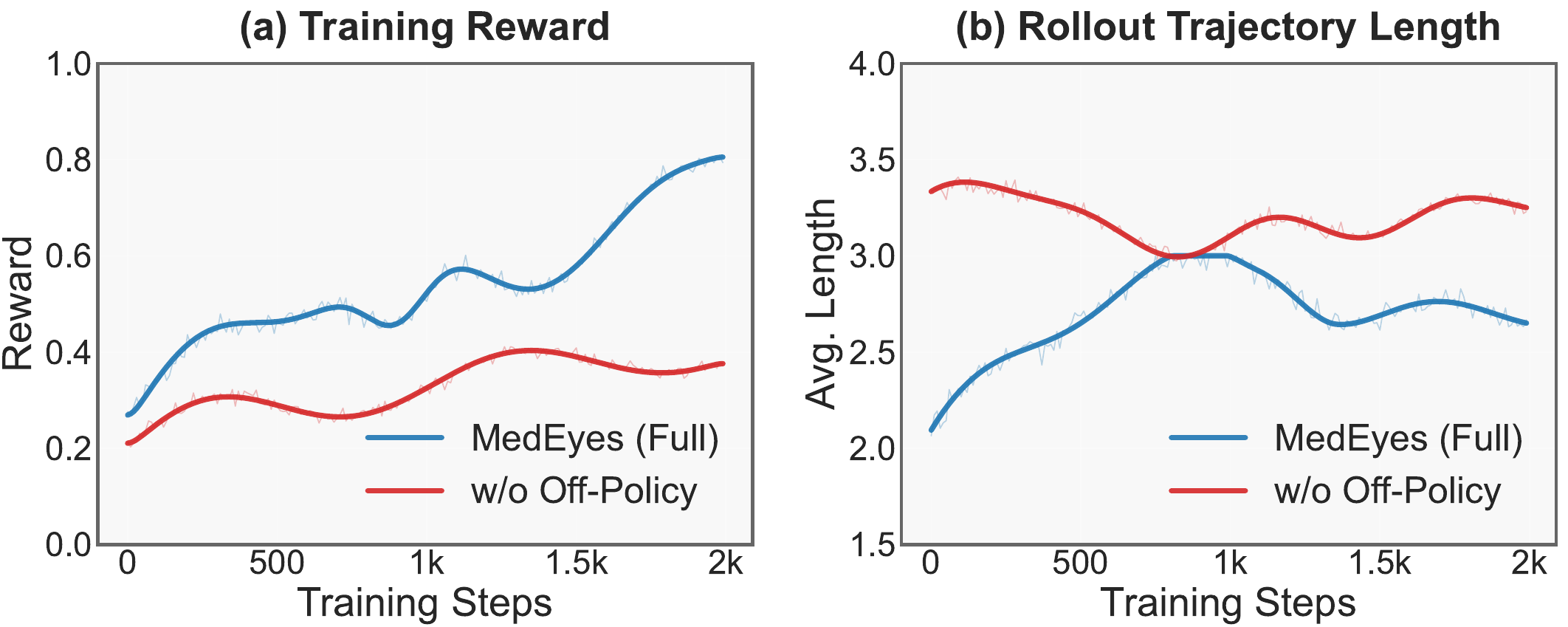}
	\caption{Training dynamics of MedEyes. (a) Reward progression highlighting the effectiveness of off-policy expert guidance. (b) Trajectory length showing exploration-efficiency transition in multi-round visual reasoning.}
  \label{training_process}
\end{figure}

%
Fig.~\ref{training_process} illustrates the training dynamics of MedEyes. The reward curve in (a) shows a steady increase during the training process, with the most significant improvement between steps 200 and 800, followed by stabilization. Compared to unguided training without off-policy trajectories, this validates the effectiveness of our dual-stream GRPO optimization in learning when to invoke visual tools, reflecting a balance between internal knowledge and external search. The average trajectory length (b) presents an interesting pattern: initially increasing from 2.1 steps to approximately 3.0 steps (exploration phase), then gradually decreasing to 2.6 steps (efficiency phase). This trajectory compression indicates that MedEyes learned to generate more concise reasoning chains while maintaining accuracy, directly reflecting the model's internalization of when visual grounding is critical and when internal knowledge is sufficiently adequate.

\begin{table}[t]
\centering
\resizebox{0.475\textwidth}{!}{
\begin{tabular}{>{\raggedright\arraybackslash}p{2.78cm}|c|cc|c}
\toprule[1pt]
\textbf{Components}
& \textbf{VQA-RAD} & \textbf{SLAKE} & \textbf{PathVQA} & \textbf{Average}\\
\midrule[0.6pt]
\textbf{MedEyes (Full)} & \textbf{70.7}$_{\pm1.2}$ & \textbf{79.1}$_{\pm0.9}$ & \textbf{64.8}$_{\pm1.3}$ & \textbf{71.5}\\
\midrule[0.6pt]
\multicolumn{5}{l}{\textit{Main Component Ablation}} \\
\xmark~GRN & 62.4$_{\pm1.4}$ & 69.8$_{\pm1.1}$ & 56.2$_{\pm1.5}$ & 62.8 \\
\xmark~CVS & 65.3$_{\pm1.3}$ & 73.5$_{\pm1.0}$ & 59.1$_{\pm1.4}$ & 66.0\\
\xmark~Off-policy & 61.2$_{\pm1.5}$ & 67.4$_{\pm1.2}$ & 54.3$_{\pm1.6}$ & 61.0\\
\midrule[0.7pt]
\multicolumn{5}{l}{\textit{GRN Design Variants}} \\
Scanning-only & 66.8$_{\pm1.2}$ & 74.2$_{\pm0.9}$ & 58.7$_{\pm1.3}$ & 66.6\\
Drilling-only & 64.5$_{\pm1.3}$ & 71.9$_{\pm1.0}$ & 60.3$_{\pm1.2}$ & 65.6\\
\bottomrule[1pt]
\end{tabular}
}
\caption{Ablation study of core components on medical VQA benchmarks. All values are reported with 95\% confidence intervals. GRN: Gaze-guided Reasoning Navigator, CVS: Confidence Value Sampler.}
\label{tab:core_ablation}
\end{table}

\begin{table}[t]
\centering
\resizebox{\columnwidth}{!}{
\begin{tabular}{lc|ccc|c}
\toprule[0.9pt]
\textbf{Config.} & \textbf{Value} & \textbf{VQA-RAD} & \textbf{SLAKE} & \textbf{PathVQA} & \textbf{Average} \\
\midrule
\multirow{4}{*}{\rotatebox{0}{\textbf{$N_{\text{expert}}$}}} 
& 2 & 64.2 & 71.8 & 57.3 & 64.4 \\
& 4 & 66.8 & 74.5 & 59.8 & 67.0 \\
& \cellcolor{gray!20}\textbf{6} 
& \cellcolor{gray!20}\textbf{70.7} 
& \cellcolor{gray!20}\textbf{79.1} 
& \cellcolor{gray!20}\textbf{64.8} 
& \cellcolor{gray!20}\textbf{71.5} \\
& 8 & 70.3 & 78.7 & 64.3 & 71.1 \\
\midrule
\multirow{4}{*}{\rotatebox{0}{\textbf{$T_{\text{max}}$}}} 
& 2 steps & 70.5 & 78.8 & 64.3 & 71.2 \\
& \cellcolor{gray!20}\textbf{3 steps} 
& \cellcolor{gray!20}\textbf{70.7} 
& \cellcolor{gray!20}\textbf{79.1} 
& \cellcolor{gray!20}\textbf{64.8} 
& \cellcolor{gray!20}\textbf{71.5} \\
& 4 steps & 70.6 & 78.2 & 63.9 & 70.9 \\
& 5 steps & 68.1 & 76.5 & 62.4 & 69.0 \\
\bottomrule[0.9pt]
\end{tabular}
}
\caption{Analysis of expert trajectory count and length on model performance. Gray background and bold text indicate the selected and optimal configurations, respectively.}
\label{tab:trajectory_ablation}
\end{table}

\subsection{Ablation Studies}

\subsubsection{Core Component Analysis}

Table~\ref{tab:core_ablation} validates our framework's core mechanisms through systematic ablation. GRN removal results in an 8.7\% performance drop, confirming that scanning-drilling strategy successfully replicates expert diagnostic workflows from systematic region identification to recursive pathological analysis. Single-mode variants demonstrate incomplete reasoning: scanning-only fails on fine-grained tasks while drilling-only lacks systematic exploration, proving both modes essential for complete medical visual reasoning. CVS removal leads to a 5.5\% degradation, validating its effectiveness in generating credible exploration trajectories. Most critically, off-policy learning removal yields 10.5\% degradation, directly confirming that expert trajectories provide indispensable cognitive anchors for breaking through autonomous exploration limitations.

\begin{table}[t]
\centering
\resizebox{\columnwidth}{!}{
\begin{tabular}{lc|ccc|c}
\toprule
\multicolumn{2}{c|}{\textbf{Config.}} & \textbf{VQA-RAD} & \textbf{SLAKE} & \textbf{PathVQA} & \textbf{Average} \\
\midrule
\multicolumn{6}{l}{\textit{Reward Weight Configuration ($\lambda_{\text{acc}}$/$\lambda_{\text{grammar}}$/$\lambda_{\text{div}}$)}} \\
& 1.0/0.0/0.0 & 68.4 & 76.2 & 62.1 & 68.9 \\
& 0.8/0.2/0.0 & 69.5 & 77.8 & 63.4 & 70.2 \\
& 0.8/0.1/0.1 & 69.9 & 78.3 & 63.8 & 70.7 \\
\rowcolor{gray!20}
& \textbf{0.7/0.2/0.1} & \textbf{70.7} & \textbf{79.1} & \textbf{64.8} & \textbf{71.5} \\
& 0.6/0.2/0.2 & 69.8 & 78.5 & 64.2 & 70.8 \\
& 0.5/0.3/0.2 & 68.3 & 77.1 & 63.0 & 69.5 \\
\bottomrule[1pt]
\end{tabular}
}
\caption{Effect of our reward function design. Configurations selected for final use are marked with gray shading, and optimal values are shown in bold.}
\label{tab:training_dynamics}
\end{table}


\subsubsection{Off-Policy Trajectory Configuration}

Table~\ref{tab:trajectory_ablation} offers important insights into how medical reasoning patterns are captured. Increasing the number of expert trajectories leads to substantial performance improvements from 2 to 6 trajectories, after which further increases provide only marginal gains, suggesting that 6 trajectories are sufficient to encode the diversity of expert diagnostic strategies. Likewise, a reasoning sequence length of 3 steps achieves the best performance, whereas longer sequences tend to accumulate errors, resulting in decreased accuracy. These findings highlight the balance between trajectory quantity and sequence length in effectively modeling expert reasoning.


\subsubsection{Reward Function Design}

Table~\ref{tab:training_dynamics} presents a detailed multi-component reward analysis, showing that optimizing for accuracy alone fails to capture structured diagnostic reasoning. Incorporating both grammar and diversity produces coherent reasoning–action–perception cycles. Grammar enforces strict syntactic grounding, while diversity promotes broader and more comprehensive exploration of diagnostically relevant regions. The optimal rewards configuration further encourages thorough and fully comprehensive exploration of diagnostically relevant regions.

\subsubsection{Off-Policy Sampling Analysis}

Table~\ref{tab:offpolicy_analysis} analyzes the impact of different off-policy trajectory sources. We compare random sampling, external trajectories from DeepSeek-R1, and historical replay, which selects 8 samples based on either recency (most recent) or reward (highest-reward). Results show that while all alternatives consistently demonstrate the benefits of off-policy learning, trajectory quality remains crucial for achieving optimal performance.

\begin{table}[t]
\centering
\resizebox{0.475\textwidth}{!}{
\begin{tabular}{>{\raggedright\arraybackslash}p{2.78cm}|c|cc|c}
\toprule[1pt]
\textbf{Strategy/Config} & \textbf{VQA-RAD} & \textbf{SLAKE} & \textbf{PathVQA} & \textbf{Average}\\
\midrule[0.6pt]
\textbf{GRN+CVS} & \textbf{70.7} & \textbf{79.1} & \textbf{64.8} & \textbf{71.5}\\
\midrule[0.7pt]
\multicolumn{5}{l}{\textit{Baseline Sampling Methods}} \\
Random & 58.3 & 65.7 & 52.1 & 58.7 \\
DeepSeek-R1 & 65.5 & 71.2 & 53.9 & 63.9 \\
\midrule[0.7pt]
\multicolumn{5}{l}{\textit{Historical Replay Strategies}} \\
Recency-based & 63.2 & 70.8 & 56.4 & 63.5 \\
Reward-oriented & 65.1 & 72.9 & 58.7 & 65.6 \\
\bottomrule[1pt]
\end{tabular}
}
\caption{Off-policy trajectory quality and sampling strategy analysis.  Gray shading indicates selected configurations.}
\label{tab:offpolicy_analysis}
\end{table}

\begin{figure}[t]
  \includegraphics[width=0.475\textwidth]{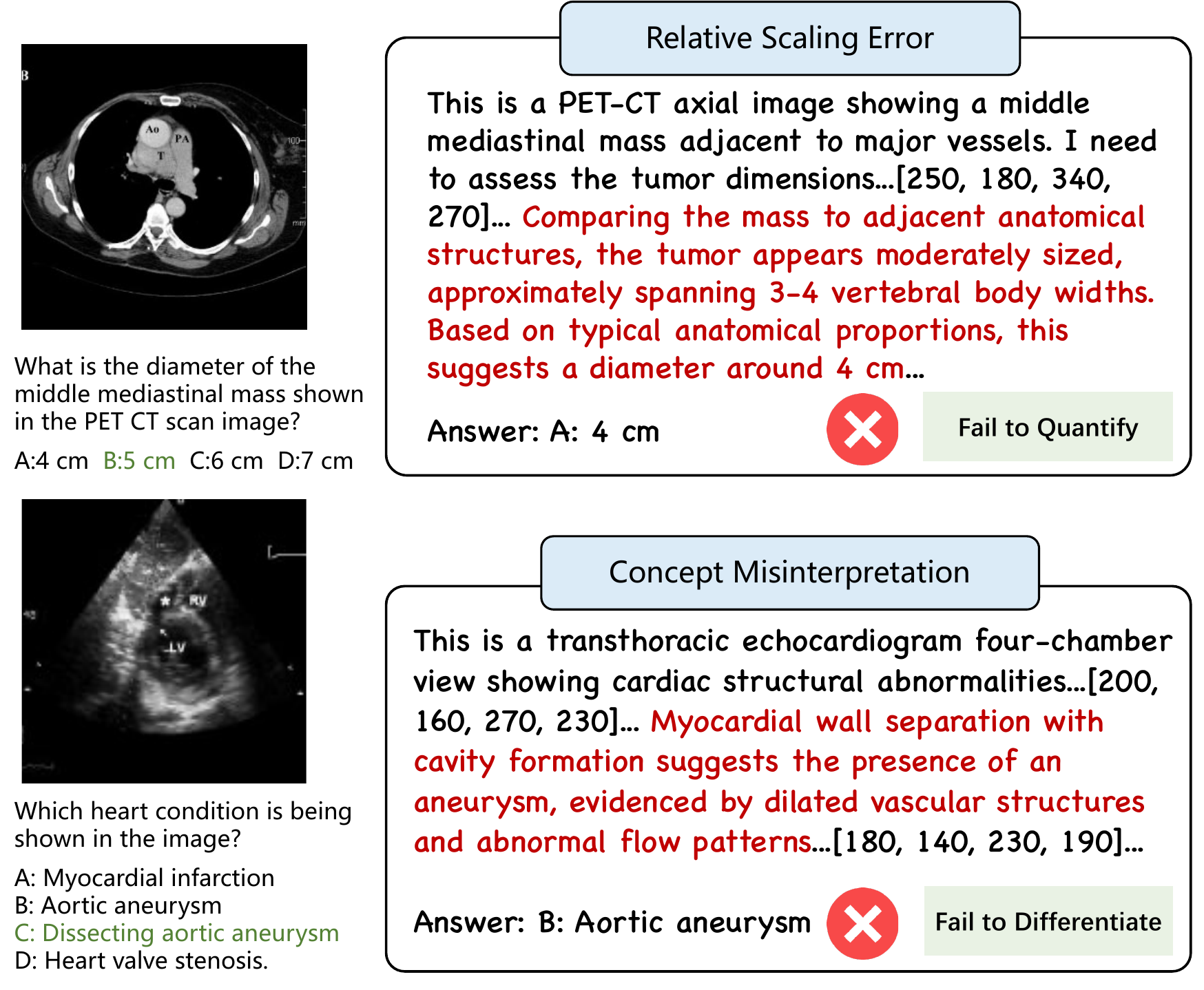}
	\caption{Failure cases analysis. quantitative measurement errors in tumor sizing (top) and pathological concept misinterpretation in ultrasound aneurysm identification (bottom).}
	\label{fig:case_analysis_error}
	\end{figure}

\subsection{Failure Case Study}

Fig. \ref{fig:case_analysis_error} demonstrates two representative error scenarios. Quantitative measurement tasks suffer from flawed relative scaling that substitutes anatomical proportion estimation for pixel-to-centimeter calibration, introducing variance incompatible with clinical requirements. Additionally, MedEyes faces challenges in fine-grained concept differentiation, occasionally conflating pathological subtypes with shared morphological features, such as ``aneurysmal changes" versus ``dissecting aneurysms". These limitations underscore the necessity of richer tools and deeper expertise understanding for future medical agents.

\section{Conclusion}

This paper presents MedEyes, a novel medical visual Chain-of-Thought (CoT) framework that enables diagnostic evolution through dynamic focus learning. By combining the scanning–drilling dual-mode exploration of a gaze-guided reasoning navigator with end-to-end RL using dual-stream GRPO, MedEyes effectively replicates the progressive visual focusing process of expert clinicians. Experimental results show that analysis of MedEyes’ dynamic focus trajectories reveals a clear shift from casual exploration to efficient visual attention, and that the method outperforms existing RL approaches on multiple medical VQA benchmarks. This visual reasoning framework offers a novel technical pathway toward building agent-driven medical systems.

\section{Acknowledgments}
This research was partially supported by grants from the National Natural Science Foundation of China (Grants No. 62472157, No. 62202158, No. 62206089), and the Science and Technology Innovation Program of Hunan Province (Grant No. 2023RC3098).


\bibliographystyle{plain}  
\bibliography{main}   

\end{document}